\documentclass[conference]{IEEEtran}
\usepackage{hyperref}

\ifCLASSINFOpdf
   \usepackage[pdftex]{graphicx}
 
\else
 
\fi
\pdfoutput=1

\begin{document}
%
% paper title
% can use linebreaks \\ within to get better formatting as desired
\title{A Discourse-based Approach in Text-based Machine Translation}

% author names and affiliations
% use a multiple column layout for up to three different
% affiliations
\author{\IEEEauthorblockN{Sana Ullah$^{\Psi}$ , M.A. Khan$^{+}$, and Kyung Sup Kwak$^{\Psi}$}
\IEEEauthorblockA{$^{\Psi}$ Graduate School of IT and Telecommunications, Inha University\\
253 Yonghyun-Dong, Nam-Gu, Incheon 402-751, South Korea.\\
$^{+}$ Department of Computer Science \\
University of Peshawar, Pakistan\\
Email: sanajcs@hotmail.com, abid\textunderscore khan1961@yahoo.com, kskwak@inha.ac.kr}}% conference papers do not typically use \thanks and this command
% is locked out in conference mode. If really needed, such as for
% the acknowledgment of grants, issue a \IEEEoverridecommandlockouts
% after \documentclass

% for over three affiliations, or if they all won't fit within the width
% of the page, use this alternative format:
% 
%\author{\IEEEauthorblockN{Michael Shell\IEEEauthorrefmark{1},
%Homer Simpson\IEEEauthorrefmark{2},
%James Kirk\IEEEauthorrefmark{3}, 
%Montgomery Scott\IEEEauthorrefmark{3} and
%Eldon Tyrell\IEEEauthorrefmark{4}}
%\IEEEauthorblockA{\IEEEauthorrefmark{1}School of Electrical and Computer Engineering\\
%Georgia Institute of Technology,
%Atlanta, Georgia 30332--0250\\ Email: see http://www.michaelshell.org/contact.html}
%\IEEEauthorblockA{\IEEEauthorrefmark{2}Twentieth Century Fox, Springfield, USA\\
%Email: homer@thesimpsons.com}
%\IEEEauthorblockA{\IEEEauthorrefmark{3}Starfleet Academy, San Francisco, California 96678-2391\\
%Telephone: (800) 555--1212, Fax: (888) 555--1212}
%\IEEEauthorblockA{\IEEEauthorrefmark{4}Tyrell Inc., 123 Replicant Street, Los Angeles, California 90210--4321}}

% use for special paper notices
%\IEEEspecialpapernotice{(Invited Paper)}

% make the title area
\maketitle

\begin{abstract}
%\boldmath
This paper presents a theoretical research based approach to ellipsis resolution in machine translation. The formula of discourse is applied in order to resolve ellipses. The validity of the discourse formula is analyzed by applying it to the real world text, i.e., newspaper fragments. The source text is converted into mono-sentential discourses where complex discourses require further dissection either directly into primitive discourses or first into compound discourses and later into primitive ones. The procedure of dissection needs further improvement, i.e., discovering as many primitive discourse forms as possible. An attempt has been made to investigate new primitive discourses or patterns from the given text.
\end{abstract}
% IEEEtran.cls defaults to using nonbold math in the Abstract.
% This preserves the distinction between vectors and scalars. However,
% if the conference you are submitting to favors bold math in the abstract,
% then you can use LaTeX's standard command \boldmath at the very start
% of the abstract to achieve this. Many IEEE journals/conferences frown on
% math in the abstract anyway.

% no keywords

% For peer review papers, you can put extra information on the cover
% page as needed:
% \ifCLASSOPTIONpeerreview
% \begin{center} \bfseries EDICS Category: 3-BBND \end{center}
% \fi
%
% For peerreview papers, this IEEEtran command inserts a page break and
% creates the second title. It will be ignored for other modes.
\IEEEpeerreviewmaketitle

\section{Introduction}
% no \IEEEPARstart
A text is not adequately translated until it is considered as part of a discourse. Discourses are linguistic units composed of several sentences. The term discourse is coined by Zellig Harris in 1952. Informally and intuitively, a discourse is a connected piece of text or spoken language of more than one sentence spoken by one or more speakers \cite{1}, \cite{2}. Discourse has been taken up in a variety of disciplines. However in our work, the term discourse is used in context of machine translation and a Discourse Unit (DU)\footnote{DU is an atomic utterance that has no reference beyond its boundaries} is taken as a unit of analysis. 

In this paper, a discourse-based approach is used to resolve anaphoric and cataphoric\footnote{It finds its consequent in the subsequent text}ambiguities. The source text is converted into mono-sentential (primitive discourses) where complex discourses require further dissection either directly into primitive discourses or first into compound discourses and later into primitive ones. An attempt has been made to (a) investigate as many primitive discourses as possible and (b) finding ways of splitting compound and complex discourses each into discourses having forms which belong to the existing set of primitive discourse forms. The discourse formula is applied to the newspaper fragments.     

\section{Dissection of complex and compound discourses into primitive ones}
We have applied discourse-based approach to several newspaper fragments theoretically and solved various anaphoric and cataphoric ambiguities. The complex and compound discourses are dissected into primitive discourses. In order to understand the concept of this paper, consider an example from our experiments:

\emph{[Around 3,500 Swedes are still missing in Thailand, a week after tidal waves struck the country's coastline, with 60 Swedes confirmed dead, the foreign ministry said Sunday]. [The ministry said, it had managed to locate the missing tourists and struck their names off the list, but new names were being added all the time].} 

There are two complex discourses in the above article enclosed by square brackets. After anaphora and cataphora resolution, complex discourses have to be dissected either into compound discourses and then later into primitive ones, OR Complex discourses are directly dissected into primitive discourses. The first complex discourse is directly dissected into a set of primitive discourses. Table \ref{tab:1} contains a set of primitive discourses compared with their generalized patterns

\begin{table}[!t]
%% increase table row spacing, adjust to taste
\renewcommand{\arraystretch}{1.3}
% if using array.sty, it might be a good idea to tweak the value of
% \extrarowheight as needed to properly center the text within the cells
\caption{Primitive discourses and Generalized patterns}
\label{tab:1}
\centering
%% Some packages, such as MDW tools, offer better commands for making tables
%% than the plain LaTeX2e tabular which is used here.
\begin{tabular}{|c|c|}
\hline
Primitive Discourses & Generalized Patterns\\
\hline
1- Swedes are 3,500 & A are B\\
\hline
2- Swedes are still missing & A are B C\\
\hline
3- Swedes are in Thailand &	A are in B\\
\hline
4- Waves were tidal &	A were B\\
\hline
5- Waves struck coastline of the country &	A B C of the D\\
\hline
6- Missing after a week &	A B a C \\
\hline
7- Swedes are 60 &	A are B \\
\hline
8- Swedes confirmed dead &	A B C\\
\hline
9- The foreign ministry said on Sunday &	The A B C on D\\
\hline
\end{tabular}
\end{table}

The generalized discourses can be used to generate hundreds of sentences having the same pattern otherwise. For instance, the second discourse ,i.e., A are B C is also valid for Children are playing game. After investigating a substantial number of generalized primitive discourses formats, the system would match the patterns of new incoming primitive discourses with the generalized discourses already stored in the machine translation system and would reject an incomplete text ,i.e., syntactically incomplete. However, it is hard to believe that a system would reject an incomplete sentence at a reasonable accuracy. 

Compound and complex discourses are dissected into discourses having forms, which belong to the existing set of primitive discourse forms. For example, the pattern A are B belongs to the existing set of primitive discourses, already 
appeared as the first generalized discourse. A list called $L^{*}$, which contains information regarding various articles, auxiliary verbs, copula verbs and preposition, is used during our theoretical experiments. It has solved the problem of redundant predicates upto great extent \cite{3}.

The second complex discourse is first converted into two compound discourses and then later into primitive discourses. After anaphora resolution, the complex discourse can be represented as: 

\emph{[The ministry said, ministry had managed to locate missing tourists and struck names of \underline{missing tourists} off the list, but new names of \underline{missing tourists} were being added \underline{throughout}].} 

Now, there are two compound discourses in the post resolution stage. 

(a) Ministry had managed to locate missing tourists and struck names of missing tourists off the list.

(b)	But new names of missing tourists were being added all the time. 

Table \ref{tab:2} contains a set of primitive discourses resulted from the compound discourse (a) and Table \ref{tab:3} contains a set of primitive discourses resulted from the compound discourse (b). 

\begin{table}[!t]
%% increase table row spacing, adjust to taste
\renewcommand{\arraystretch}{1.3}
% if using array.sty, it might be a good idea to tweak the value of
% \extrarowheight as needed to properly center the text within the cells
\caption{Primitive discourses and Generalized patterns}
\label{tab:2}
\centering
%% Some packages, such as MDW tools, offer better commands for making tables
%% than the plain LaTeX2e tabular which is used here.
\begin{tabular}{|c|c|}
\hline
Primitive Discourses & Generalized Patterns\\
\hline
10- Tourists are missing &	A are B \\
\hline
11- Ministry had managed &	A had B\\
\hline
12- Tourists are located &	A are B \\
\hline
13- Tourists are missing &	A are B\\
\hline
14- Struck names off the list &	A B C the D\\
\hline
\end{tabular}
\end{table}

\begin{table}[!t]
%% increase table row spacing, adjust to taste
\renewcommand{\arraystretch}{1.3}
% if using array.sty, it might be a good idea to tweak the value of
% \extrarowheight as needed to properly center the text within the cells
\caption{Primitive discourses and Generalized patterns}
\label{tab:3}
\centering
%% Some packages, such as MDW tools, offer better commands for making tables
%% than the plain LaTeX2e tabular which is used here.
\begin{tabular}{|c|c|}
\hline
Primitive Discourses & Generalized Patterns\\
\hline
15- Names are new	& A are B\\
\hline
16- Tourists are missing &	A are B\\
\hline
17- Names were being added throughout &	A were B C D\\
\hline
\end{tabular}
\end{table}

\section{Experiments and Evaluation}

\begin{figure}[!t]
\centering
\includegraphics[width=3in]{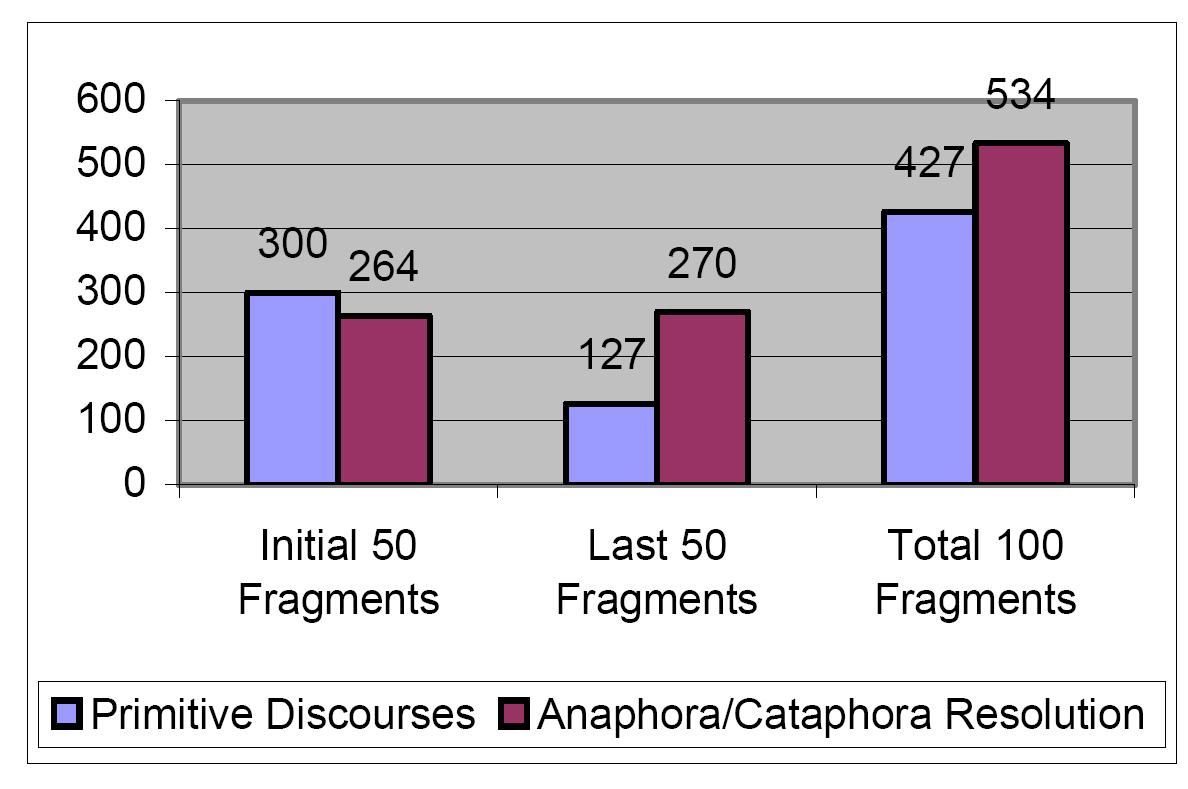}
% where an .eps filename suffix will be assumed under latex, 
% and a .pdf suffix will be assumed for pdflatex; or what has been declared
% via \DeclareGraphicsExtensions.
\caption{Dissection of complex and compound discourses into primitive ones}
\label{fig:1}
\end{figure}

The discourse formula, i.e., dissection of complex and compound discourses into primitive ones, is applied manually to 100 newspaper fragments. Our aim was to investigate as many primitive discourses as possible and to resolve anaphoric and cataphoric ambiguities during dissection. During our experiments we have investigated approximately 427 new primitive discourses having completely new patterns. More than 534 anaphoric/cataphoric ambiguities were resolved. Note that during our experiment we got considerable number of redundant predicates, i.e., predicates having the same pattern. For example, we got the pattern 'A are B' approximately more than 100 times, but it was considered only once. The results are given in Fig. \ref{fig:1}. The figure shows that the discourse formula being applied to the initial 50 newspaper fragments resulted into approximately 300 primitive discourses and the resolution of 264 anaphoric/cataphoric ambiguities. In the last 50 fragments, the numbers of new primitive discourses have been decreased up to 127 primitive discourses, while the resolution of anaphoric/cataphoric ambiguities remained the same. We have concluded that as we go further to investigate new primitive discourses, the number of new primitive discourses decrease accordingly.  
% An example of a floating figure using the graphicx package.
% Note that \label must occur AFTER (or within) \caption.
% For figures, \caption should occur after the \includegraphics.
% Note that IEEEtran v1.7 and later has special internal code that
% is designed to preserve the operation of \label within \caption
% even when the captionsoff option is in effect. However, because
% of issues like this, it may be the safest practice to put all your
% \label just after \caption rather than within \caption{}.
%
% Reminder: the "draftcls" or "draftclsnofoot", not "draft", class
% option should be used if it is desired that the figures are to be
% displayed while in draft mode.
%

% Note that IEEE typically puts floats only at the top, even when this
% results in a large percentage of a column being occupied by floats.

\section{Conclusions} 
A discourse-based approach in text-based machine translation was the main focus of this paper where DU was taken as a unit of analysis. The source text was dissected into multiple discourses, i.e., mono-sentential or poly-sentential. After various ambiguities were resolved, the complex discourses were dissected into mono-sentential discourses. The mono-sentential discourse can be easily translated into its corresponding target language, using different implementation tools such as PROLOG and LISP. However, choosing PROLOG could be the best implementation tool owing to the fact that the resultant primitive discourses are easily represented into their corresponding prolog statements, which is later used for translation.

This paper presented the mechanism of how a source text is dissected into mono sentential discourses, for translation purposes. However, after achieving the translation of mono sentential discourses there is a need to rearrange the mono sentential discourses back into complex discourses of the target language. This could be the reverse mechanism of dissecting the source language. In future our aim is to improve this work by implementing the resultant primitive discourses using appropriate language, translate the discourses into corresponding target language, and then rearrange the translated primitive discourses into complex discourses of the target language.     

\section*{Acknowledgement}

This research is supported by the MIC (Ministry of Information and Communication) South Korea, under ITRC (Information Technology Research Center) support program supervised by II TA  (Institute of Information Technology Advancement).

\begin{thebibliography}{spphys}

\bibitem{1}
Khan, M.A., Text Based Machine Translation, Department of Computer Science, University of Peshawar, 1995.
\bibitem{2}
Sidner, C. L., Comprehension of Definite Anaphora. Readings in NLP Eds. Bibel, W. and Jorrand, pp277-313, 1987. 
\bibitem{3} 
Ullah, S., MSc Research Thesis: Ellipsis Resolution in Machine Translation, Department of Computer Science, University of Peshawar, 2006.
\end{thebibliography}
\end{document}